\documentclass[letterpaper, 10 pt, conference]{ieeeconf}
\IEEEoverridecommandlockouts    
\usepackage{hyperref}
\usepackage{soul}
\usepackage{makecell}

\usepackage{threeparttable}

\usepackage{caption}
\hypersetup{
    colorlinks=true,
    linkcolor=magenta,
    filecolor=magenta,      
    urlcolor=magenta,
}

\usepackage{graphics}           
\usepackage{times}              
\usepackage{amsmath}            
\usepackage{amssymb}            
\usepackage{graphicx}
\usepackage{algorithm}
\usepackage[noend]{algpseudocode}
\usepackage{booktabs}
\usepackage{color}
\definecolor{instructioncolor}{rgb}{.5,.5,.5}

\usepackage[font=small]{caption}

\def\figref#1{Fig.~\ref{#1}}

\def\eqref#1{Eq.~(\ref{#1})}

\makeatletter
\usepackage{xspace}
\DeclareRobustCommand\onedot{\futurelet\@let@token\@onedot}
\def\@onedot{\ifx\@let@token.\else.\null\fi\xspace}

\makeatother

\usepackage{array}
\newcolumntype{L}[1]{>{\raggedright\let\newline\\\arraybackslash\hspace{0pt}}m{#1}}
\newcolumntype{C}[1]{>{\centering\let\newline\\\arraybackslash\hspace{0pt}}m{#1}}
\newcolumntype{R}[1]{>{\raggedleft\let\newline\\\arraybackslash\hspace{0pt}}m{#1}}

\usepackage{bm}
\usepackage{multirow}
\usepackage{rotating}  

\title{\LARGE \bf LT-Gaussian: Long-Term Map Update Using 3D Gaussian Splatting \\ for Autonomous Driving}

\author{Luqi Cheng \and Zhangshuo Qi \and Zijie Zhou \and Chao Lu \and Guangming Xiong$^{*}$ 
  \thanks{
  This work was supported by the National Natural Science Foundation of China under Grant 52372404.
  }
  \thanks{
  Luqi Cheng, Zhangshuo Qi, Zijie Zhou, Chao Lu and Guangming Xiong are with Beijing Institute of Technology, Beijing, 100081, China
  }
  \thanks{
  $^{*}$Corresponding author (xiongguangming@bit.edu.cn)
  }
}

\begin{document}
\maketitle
\pagestyle{empty}  
\thispagestyle{empty} 

\IEEEpeerreviewmaketitle
\thispagestyle{empty}
\pagestyle{empty}

\begin{abstract}

Maps play an important role in autonomous driving systems. The recently proposed 3D Gaussian Splatting (3D-GS) produces rendering-quality explicit scene reconstruction results, demonstrating the potential for map construction in autonomous driving scenarios. However, because of the time and computational costs involved in generating Gaussian scenes, how to update the map becomes a significant challenge. In this paper, we propose LT-Gaussian, a map update method for 3D-GS-based maps. LT-Gaussian consists of three main components: Multimodal Gaussian Splatting, Structural Change Detection Module, and Gaussian-Map Update Module. Firstly, the Gaussian map of the old scene is generated using our proposed Multimodal Gaussian Splatting. Subsequently, during the map update process, we compare the outdated Gaussian map with the current LiDAR data stream to identify structural changes. Finally, we perform targeted updates to the Gaussian-map to generate an up-to-date map. We establish a benchmark for map updating on the nuScenes dataset to quantitatively evaluate our method. The experimental results show that LT-Gaussian can effectively and efficiently update the Gaussian-map, handling common environmental changes in autonomous driving scenarios. Furthermore, by taking full advantage of information from both new and old scenes, LT-Gaussian is able to produce higher quality reconstruction results compared to map update strategies that reconstruct maps from scratch. Our open-source code is available at \url{https://github.com/ChengLuqi/LT-gaussian}.\\

\end{abstract}

\section{Introduction}
\label{sec:intro}

In many autonomous driving frameworks, maps serve as a fundamental component that provides the perceptual basis for downstream tasks~\cite{kim2023perception,li2022hdmapnet}. For example, point cloud maps can provide place recognition methods with priori localization information~\cite{qi2024gsprmultimodalplacerecognition,wang2024cctnet} and enhance object detection methods with contextual information~\cite{ravi2018real,zhang2023radar}, thereby facilitating more informed driving decisions. 
When used in the real world, environmental maps~\cite{xu2022fast,campos2021orb} primarily face the challenge of map updates. Maps must be reconstructed on a regular basis to ensure that the fine-grained scene details are up-to-date. However, large-scale map reconstruction involves significant economic and computational costs.
As a result, in autonomous driving scenarios, where changes in environmental factors and road infrastructure occur frequently, maps are often unable to meet the demand for up-to-date information from autonomous vehicles~\cite{ni2023improved,stefanini2023safe}.

In recent years, map update methods based on structural change detection~\cite{kim2022lt,stefanini2022efficient} have shown potential for efficient map update. These methods compare the current sensor data stream with the outdated map, to identify areas where structural changes have occurred. Subsequently, the areas that have undergone structural changes in the map are selectively updated in order to obtain an up-to-date map. These methods reduce the cost of map reconstruction while effectively utilizing the prior information provided by outdated maps.

\begin{figure}
  \centering
  \includegraphics[width=1\linewidth]{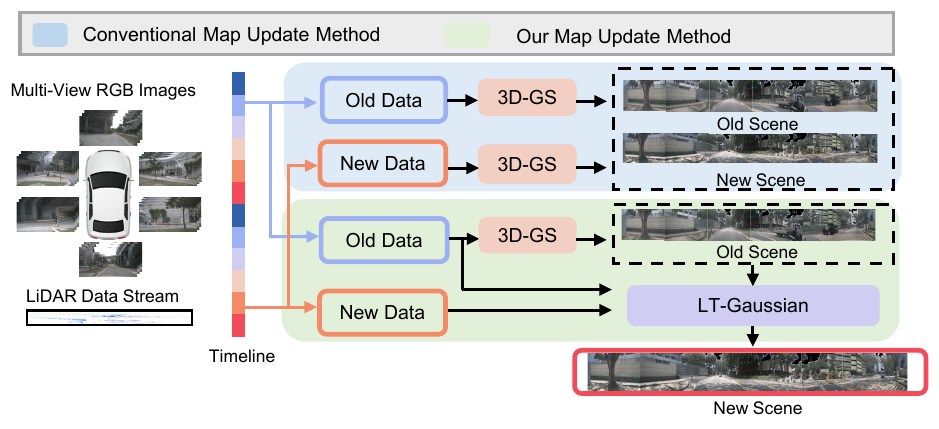}
  \caption{Environmental features of the same place change over time. By detecting structural changes between old and new scenes, LT-Gaussian enables efficient updating of high-resolution maps created by 3D Gaussian Splatting.}
  \label{fig:introduction}
  \vspace{-0.6cm}
\end{figure}

Recently, 3D Gaussian Splatting (3D-GS)~\cite{kerbl20233d} has been proposed to construct an explicit scene representation using 3D Gaussians, capable of generating rendering-level 3D scene reconstruction results. Recent work~\cite{yan2024street} has extended 3D-GS from bounded scenes to outdoor environments, enabling 3D-GS-based environmental maps (Gaussian-map) for autonomous driving. However, the time and computational cost of map updates become particularly apparent, when applying 3D-GS to autonomous driving scenarios. In particular, training a single scene can take tens of minutes or even hours~\cite{yan2024street, zhou2024drivinggaussian}. In addition, how to perform map updates on Gaussian-map remains an open problem, further increasing the difficulty of applying 3D-GS in autonomous driving.

In this paper, we propose LT-Gaussian, a map update method for 3D-GS-based maps, as shown in ~\figref{fig:introduction}. Inspired by~\cite{kim2022lt}, we propose a structural change detection method for Gaussian-maps, enabling targeted map updates.
To the best of our knowledge, this is the first Gaussian-map update method. Compared to the map update strategy of reconstructing from scratch, it effectively leverages the prior information in the outdated Gaussian-map to produce higher quality reconstruction results. In addition, it significantly reduces the time cost of map updates, providing a viable solution for updating Gaussian-maps in autonomous driving scenarios.

In summary, our main contributions are as follows:
\begin{itemize}
\item We propose LT-Gaussian, a Gaussian-map updating method, which preserves unchanging environmental structures over long time spans and updates changing environmental structures in a timely manner.
\item We propose a structural change detection method for Gaussian-map, which detects emerging and disappearing points by comparing outdated Gaussian-map with LiDAR point clouds from the current sensor stream.
\item Extensive experiments show that our method outperforms multimodal 3D-GS baseline in terms of reconstruction quality, while significantly reducing the map update time.
\end{itemize}

\section{Related Work}
\label{sec:related}

\subsection{Environmental Map Construction}
\label{sec:scene}
Maps contain fine-grained scene details in autonomous driving scenarios and serve as the basis for map updates. Most environmental map construction methods~\cite{hu2024admap,jo2018simultaneous} have relied primarily on LiDAR, stereo vision systems, and depth cameras.
Due to the ability of cameras to provide rich semantic information at a lower cost, image-based map construction methods have seen advances~\cite{bastani2021updating,liu2024less}. However, images are susceptible to interference from changes in illumination, occlusion and dynamic objects, resulting in less stable performance in dynamic environments. 

LiDAR-based map construction methods ensure that autonomous driving systems can maintain highly accurate navigation capabilities in changing environments by capturing fine-grained structures. Several SLAM methods have made map construction possible~\cite{zhang2014loam, shan2018lego, shan2020lio}. LOAM~\cite{zhang2014loam} extracts line and plane features from point clouds for odometry and ranging via registrations. LeGO-LOAM~\cite{shan2018lego} reduces the computational load by ground segmentation and loop closure detection. 
LIO-SAM~\cite{shan2020lio} combines IMU preintegration strategy with LiDAR odometry, and uses graph optimization to improve the consistency of the map.

\begin{figure*}[ht]
  \centering
  \includegraphics[width=1\linewidth]{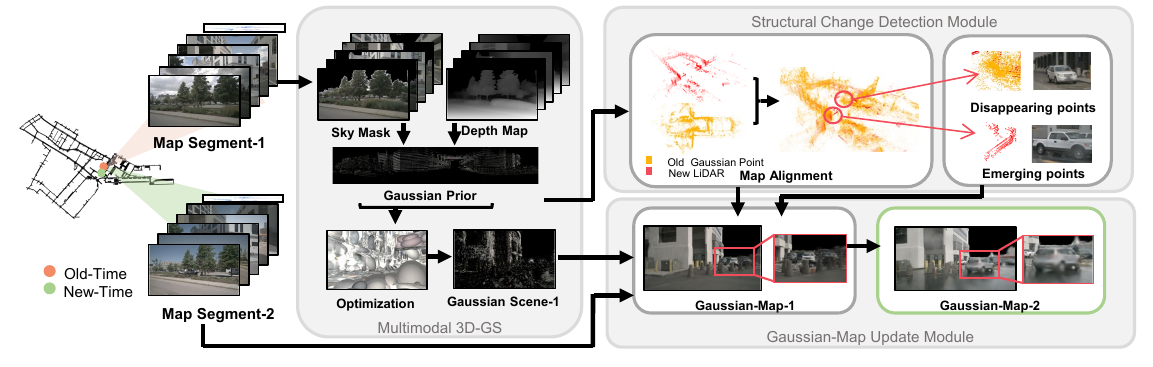}
  \caption{The overall architecture of LT-Gaussian. Images and LiDAR point clouds from Map Segment-1 (Old-Time) and Map Segment-2 (New-Time) are input. Semantically masked multimodal data is used to generate the initial Gaussian-map through iterative refinement under the supervision of depth priors. In the Gaussian-Map Update module, the initial Gaussian-map is used for registration with the LiDAR data stream from Map Segment-2, followed by structural change detection. The processed initial Gaussian-map is then used as a prior to reconstruct the new scene, achieving an efficient Gaussian-map update.}
  \label{fig:2}
  \vspace{-0.6cm}
\end{figure*}

\subsection{LiDAR-Based Map Update}
\label{sec:scene}
In autonomous driving scenarios, environmental features change over time, affecting the accuracy of the map. Therefore, it is critical to regularly update the environmental maps to provide autonomous vehicles with up-to-date environmental information.

To address this issue, Chen \textit{et al}.~\cite{chen2024slam} propose SLAM-RAMU, which integrates 3D LiDAR and IMU for lifetime SLAM and autonomous map updates. Meanwhile, Jinno \textit{et al}.~\cite{jinno20193d} used a mobile robot with LiDAR for 3D map changes in human environments, and Stefanini \textit{et al}.~\cite{stefanini2022efficient} calculate the difference between the map point cloud and the scanned point cloud and update the area with excessive difference.
To improve the efficiency of map updating, the method proposed by Kim \textit{et al}.~\cite{kim2020remove} is specifically designed to address environmental change detection and alignment errors in dynamic environments. 
By introducing intra-session and inter-session change detection, it accurately distinguishes dynamic objects and structures, resulting in more accurate and robust map updates.

Building on the advantages of the above methods, LT-Mapper~\cite{kim2022lt} updates the map by detecting structural changes in the environment, achieving efficient map update. By comparing the current LiDAR data stream to previous observations of the same location, it efficiently identifies structural changes. Then, targeted updates are performed in areas where structural change occurs.

\subsection{3D-GS-Based Environmental Map}
\label{sec:mpr}
When constructing maps for autonomous driving scenarios, unimodal systems often have shortcomings. Image-based mapping methods are susceptible to environmental changes. LiDAR-based mapping methods, while capable of accurately capturing scene structural details, suffer from a lack of semantic and texture information. Recently, the emergence of 3D-GS~\cite{kerbl20233d} has provided a new perspective for constructing environmental maps. In particular, 3D-GS not only allows explicit reconstruction of the geometric structure, but also encodes rich texture information from the images. Furthermore, compared to implicit neural reconstruction methods such as NeRF~\cite{mildenhall2021nerf}, the explicit scene representations generated by 3D-GS are more suitable for downstream tasks such as navigation and planning, showing significant potential for autonomous driving scenarios. 

While 3D-GS performs well in bounded environments, it faces challenges in autonomous driving scenarios, particularly due to scale uncertainty and overfitting of training views. To overcome these issues, Street Gaussian~\cite{yan2024street} incorporates LiDAR point cloud data to provide geometric prior and scale information. Driving Gaussian~\cite{zhou2024drivinggaussian} introduces a dynamic Gaussian-map to enable dyanmic scene reconstruction. PVG~\cite{chen2023periodic} improves the expression of the original Gaussian function to better suit urban scenarios. Huang \textit{et al}.~\cite{huang2024textit} introduces a self-supervised foreground-background decomposition using a multi-resolution hexadecimal plane, reducing the dependence on annotated data. 

However, the above methods have the drawback of long training times and high costs, making the map update strategy of reconstructing the map from scratch burdensome. Therefore, we propose a Gaussian-map update method based on structural change detection, which efficiently produces high-quality reconstruction results of the new scene based on the outdated Gaussian-map.

\section{Our Approach}
\label{sec:method}

The overview of our proposed LT-Gaussian is depicted in~\figref{fig:2}. The system consists of three main components: Multimodal Gaussian Splatting, Structural Change Detection Module and Gaussian-Map Update Module. The Gaussian-map is generated by harmonizing multi-view RGB images and LiDAR point clouds, and serves as a rendering-level map for autonomous driving scenarios. The Structural Change Detection module analyzes and compares the differences in the geometric structure between the outdated Gaussian-map and the current LiDAR data stream. This process accurately identifies the structural changes in the map over time. Based on the detected areas of structural change, the Gaussian-Map Update module effectively performs targeted map updates. By using the processed outdated Gaussian-map as a prior, accurate map updates can be achieved in a much shorter time.

\subsection{Multimodal 3D-GS-Based Map Construction}
\label{sec:multimodal_3dgs}

We propose using Multimodal Gaussian Splatting (Multimodal 3D-GS) to integrate multi-view images and LiDAR data to create rendering-quality autonomous driving scene representations. Let \(\mathcal{C} = \{c_1, c_2, \cdots, c_n\}\) be the multi-view images and \(\mathcal{L}=\{l_1, l_2, \cdots, l_m\}\) be the sequential LiDAR data. 
In autonomous driving scenarios, the sparse distribution of viewpoints makes it difficult for COLMAP~\cite{fisher2021colmap} to produce usable sparse reconstruction results. Thus, we concatenate the sequential LiDAR data $\mathcal{L}$ based on the localization results as a LiDAR submap $\mathcal{S}$, providing accurate geometric structure priors and metric depth for the Gaussian initialization process. Multi-view images $\mathcal{C}$ are then used to supervise the iterative refinement of the Gaussian attributes, in order to produce high-quality reconstruction results~\cite{kerbl20233d}.

During the optimization process, iterative optimization with the goal of rendering quality often leads to misalignment of the geometric structure. To overcome this, we use Depth Anything V2~\cite{yang2024depth} to generate per-pixel relative depth annotations for training images, providing additional geometry control.
Following~\cite{xiong2023sparsegs}, the Pearson correlation loss is used for nonmetric mean and covariance-independent depth supervision. Let \(\hat{D}\) be the estimated depth and \(D\) be the rendered depth. The Pearson correlation loss \(L_{P}\) is calculated as:
\begin{equation}
L_{P}=1 - \frac{\sum_{i = 1}^{N}(\hat{D}_i-\bar{\hat{D}})(D_i - \bar{D})}{\sqrt{\sum_{i = 1}^{N}(\hat{D}_i-\bar{\hat{D}})^2\sum_{i = 1}^{N}(D_i - \bar{D})^2}}
\end{equation}
where \(N\) is the number of pixels in the depth map, \(\bar{\hat{D}}\) is the mean of the estimated depth, and \(\bar{D}\) is the mean of the rendered depth. Under the supervision of the Pearson correlation loss, a more geometrically consistent Gaussian-map can be obtained.

In large outdoor scenes, distant landscapes that have no geometric priors, such as the sky, can lead to structural ambiguity. To address this problem, we use Mask2Former~\cite{cheng2022masked} to generate semantic segmentation results from multi-view RGB images to create masks for the sky region. 
The sky masks play a role in the iterative refinement process of the Gaussians. We fill the pixels inside the mask region with the background color of the Gaussian rasterizer to limit the generation of Gaussians that do not correspond to any geometric priors. Ultimately, the resulting Gaussian-map exhibit a scale that consistent with the real scene, and also accurately reflect the fine-grained geometric structure.

\subsection{Structural Change Detection}
\label{sec:GDG}
\begin{figure}
  \centering
  \includegraphics[width=1\linewidth]{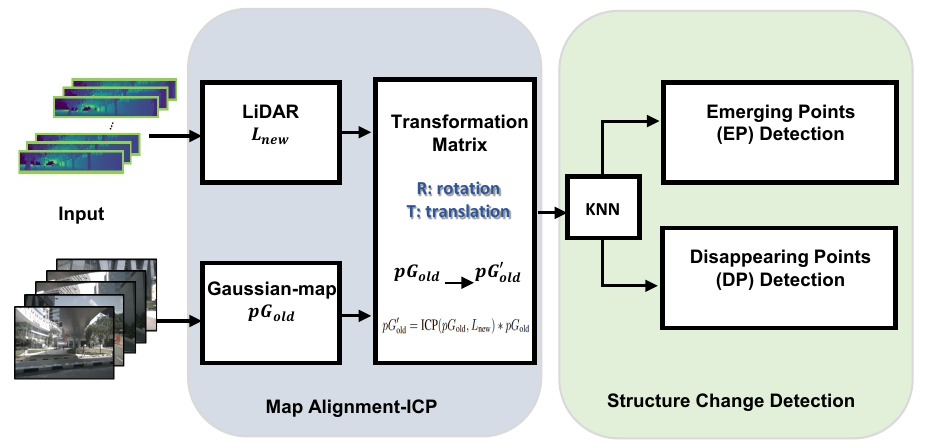}
  \caption{The pipeline of the structural change detection module. Emerging and disappearing points are selected through a two-step process of ICP-based registration and kNN-based search.}
  \label{fig:3}
  \vspace{-0.6cm}
\end{figure}

In real-world autonomous driving scenarios, the environmental features of the same location change over time. To perform targeted map updates, it is necessary to detect structural changes between the new and old scenes. As shown in~\figref{fig:3}, we compare the spatial structure of the outdated Gaussian-map with the online LiDAR data stream, to detect structural changes in the map. Firstly, Gaussian-map to LiDAR registration is applied to align the geometric structure of the scenes. Then, a search-based structural change detection method is used to identify the emerging points in the new scene and the disappearing points in the old scene.

\subsubsection{Map Alignment} Let \(\mathcal{S}=\{p_i=[x_i^\text{ls}, y_i^\text{ls}, z_i^\text{ls}]^T\in\mathbb{R}^3\}_{i = 1,\ldots,N}\) 
be a LiDAR submap and \(G=\{g_{j}=[x_j^\text{gs}, y_j^\text{gs}, z_j^\text{gs}, s_{j}, q_{j}, sh_{j}, \alpha_{j}]^T \in \mathbb{R}^{59}\}_{j = 1,\ldots,M}\) be a set of Gaussians. The coordinates \(x_i^\text{ls},y_i^\text{ls},z_i^\text{ls}\) and \(x_j^\text{gs},y_j^\text{gs},z_j^\text{gs}\) represent the positions of the \(i\)-th LiDAR point and \(j\)-th Gaussian in the Cartesian coordinate system, respectively. Apart from position, each Gaussian has attributes such as scale matrix \(s_{j}\), quaternion \(q_{j}\), SH coefficients \(sh_{j}\) and opacity \(\alpha_{j}\).

To register the Gaussians with the LiDAR points, a six-degree-of-freedom rigid body transformation with rotation \(\mathbf{R}\) and translation \(\mathbf{t}\) is applied. For each Gaussian \(g_{j}\) in \(G\), its position \(pG=[x_j^\text{gs}, y_j^\text{gs}, z_j^\text{gs}]^T\) should be as close as possible to the corresponding point \(p_i\) in \(\mathcal{S}\) after the transformation. We can find \(\mathbf{R}\) and \(\mathbf{t}\) by minimizing the following objective function:
\begin{equation}
\underset{\mathbf{R},\mathbf{t}}{\text{argmin}} \sum_{j = 1}^{M} \sum_{i = 1}^{N} \left\lVert \mathbf{R} \cdot 
\begin{bmatrix}
x_j^\text{gs} \\ 
y_j^\text{gs} \\ 
z_j^\text{gs}
\end{bmatrix} + \mathbf{t} - 
\begin{bmatrix}
x_i^\text{ls} \\ 
y_i^\text{ls} \\ 
z_i^\text{ls}
\end{bmatrix}
\right\rVert^2
\end{equation}
where \(\left\lVert\cdot\right\rVert^2\) represents the square of the Euclidean norm.

We use the Iterative Closest Point (ICP) algorithm~\cite{bouaziz2013sparse} to estimate the optimal transformation \(T_{\text{o}}\) between the Gaussian-map \(G_{\text{old}}\) and the LiDAR submap \(\mathcal{S}_{\text{new}}\). Using \(T_{\text{o}}\), we transform the position property  \(pG_{\text{old}}\) of \(G_{\text{old}}\) to \(pG_{\text{old}}'\) to align with \(\mathcal{S}_{\text{new}}\) via:
\begin{equation}
pG_{\text{old}}' = \text{ICP}(pG_{\text{old}}, \mathcal{S}_{\text{new}})*pG_{\text{old}}
\label{eq:scenario_registration}
\end{equation}
where \(\text{ICP}(\cdot)\) is the ICP registration operator.

\subsubsection{Emerging Points (EP) Detection} Based on the nature of the structural changes, we define the points that do not appear in the outdated Gaussian-map but are present in the current LiDAR submap as emerging points. We select emerging dynamic points based on the K-Nearest Neighbor (kNN) algorithm~\cite{peterson2009k}. Using the LiDAR submap \(\mathcal{S}_{\text{new}}\) as the query point set and the position property \(pG_{\text{old}}'\) of the Gaussian-map as the database point set, we can perform search-based structural change detection.

Let the nearest-neighbor threshold be $r$. Denote the number of nearest neighbors to be found by the kNN algorithm as $h$ (where $h = 10$). For a point $S_{\text{new},i}$ in $S_{\text{new}}$, we obtain a set of $h$ nearest neighbors from the Gaussian - map position property $p_{G'_{\text{old}}}$ using the kNN operation:
\begin{equation}
\mathcal{P}_{H}=kNN(S_{\text{new},i}, p_{G'_{\text{old}}}, h)
\end{equation}
where $\mathcal{P}_{H}$ denotes a point set obtained by kNN operation. 

The selection process of emerging dynamic points can be formulated as:

\begin{equation}
EP(S_{\text{new},i})=
\begin{cases}
1, & \frac{1}{h}\sum_{p_{h}\in\mathcal{P}_{H}}\|p_{h}-S_{\text{new},i}\|_2\geq r\\
0, & \frac{1}{h}\sum_{p_{h}\in\mathcal{P}_{H}}\|p_{h}-S_{\text{new},i}\|_2< r
\end{cases}
\end{equation}
where $EP$ represents the emerging points, $\|\cdot\|_2$ represents the 2-norm operation, and $kNN(.)$ represents the kNN operator.


\subsubsection{Disappearing Points (DP) Detection} Conversely, we define the points that appear in the outdated Gaussian - map but no longer exist in the current LiDAR submap as disappearing points. Similar to the emerging points detection, for a point $p_{G'_{\text{old},i}}$ in $p_{G'_{\text{old}}}$, we still use the LiDAR submap $S_{\text{new}}$ as the database point set and find a set of $h$ nearest neighbors from it using the kNN operation:
\begin{equation}
\mathcal{P}_{H}'=kNN(p_{G'_{\text{old},i}}, S_{\text{new}}, h)\
\end{equation}
where $p_{h}'\in\mathcal{P}_{H}'$. 

Given a nearest-neighbor threshold $r'$, the selection process of disappearing points can be formulated as:

\begin{equation}
DP(p_{G'_{\text{old},i}})=
\begin{cases}
1, & \frac{1}{h}\sum_{p_{h}'\in\mathcal{P}_{H}'}\|p_{h}'-p_{G'_{\text{old},i}}\|_2\geq r'\\
0, & \frac{1}{h}\sum_{p_{h}'\in\mathcal{P}_{H}'}\|p_{h}'-p_{G'_{\text{old},i}}\|_2< r'
\end{cases}
\end{equation}

where $DP$ represents the disappearing points, $\|\cdot\|_2$ represents the 2-norm operation, and $kNN(.)$ represents the kNN operator.


\subsection{Gaussian-Map Update}
\label{sec:network_training}
After detecting emerging and disappearing points in the structural change detection module, our goal is to perform targeted update on the outdated Gaussian-map. Besides aligning the \(x,y,z\) coordinates, we need to transfer additional attributes \((q_{j}, sh_{j})\) of Gaussians from the old scene to the new one.

We use the Wigner-D matrix~\cite{weinbub2018recent} to transform the anisotropic spherical harmonics. Given the rotation matrix \(\mathbf{R}_{\text{o}}\) obtained during map alignment, the spherical harmonic coefficient \(sh_{j}\) in the new scenario is calculated as:
\begin{equation}
sh_{j}^{\text{new}}=\sum_{m'}D_{mm'}^l(\mathbf{R}_{\text{o}})sh_{j}^{\text{old}}
\end{equation}
where \(D_{mm'}^l(\mathbf{R}_{\text{o}})\) is an element of the Wigner-D matrix corresponding to the rotation \(\mathbf{R}_{\text{o}}\). The parameter \(l\) represents the degree of the spherical harmonic, \(m\) and \(m'\) are the orders of the spherical harmonics before and after the rotation respectively. 

Attributes that are independent of the scene coordinate system, such as the scaling matrix \(s_{j}\) and opacity \(\alpha_{j}\), remain unchanged. Thus, we can obtain the Gaussian-map transformed to the new scene coordinate system, represented as \(G_{\text{old}}'=\{g_{j}=[pG_{\text{old}, {j}}, s_{j}, q_{j}^{\text{new}}, sh_{j}^{\text{new}}, \alpha_{j}]^T \in \mathbb{R}^{59}\}_{j = 1,\ldots,M}\).

Subsequently, we remove the disappearing points \(DP\) from the transformed Gaussian-map \(G_{\text{old}}'\). The resulting set, denoted as \(G_{\text{old}}''\), represents the intermediate Gaussian set that retains only the unchanged points. The formal definition is given by:
\begin{equation}
G_{\text{old}}''=G_{\text{old}}'\setminus DP
\end{equation}

We then add the emerging points \(EP\) to the Gaussian set \(G_{\text{old}}''\). The LiDAR submap \(\mathcal{S}_{\text{new}}\) has only the position property. To assign other features to each emerging point, we average the features of its \(e\) nearest Gaussians \(G_{\text{n}}\) in \(G_{\text{old}}''\). 
Let \(EP_{i}'\) be the emerging points assigned with feature (\(s\), \(q\), \(sh\), and \(\alpha\)). The definition of \(EP_{i}'\) is given by:
\begin{equation}
EP_{i}' = {\left[\mathcal{S}_{\text{new},i}, 
\frac{\sum\limits_{G_{\text{n},i}}s_{j}}{e}, \frac{\sum\limits_{G_{\text{n},i}}q_{j}}{e}, \frac{\sum\limits_{G_{\text{n},i}}sh_{j}}{e},  \frac{\sum\limits_{G_{\text{n},i}}\alpha_{j}}{e}\right]^{T}}
\end{equation}

We then combine \(G_{\text{old}}''\) with the emerging points with assigned features, to obtain the Gaussian prior \(G_{\text{m}}\).
\begin{equation}
G_{\text{m}}=G_{\text{old}}''\cup EP_{i}'
\end{equation}

In the process of generating the Gaussian-map \(G_{\text{new}}\) for the new scene, we use \(G_{\text{m}}\) as the initial Gaussians. By taking advantage of the rich priors contained in \(G_{\text{m}}\) from the old scene, we are able to encode the features of new scenes in a small number of refinement iterations. Ultimately, the update of the Gaussian-map is achieved efficiently.

\begin{figure}
  \centering
  \includegraphics[width=1\linewidth]{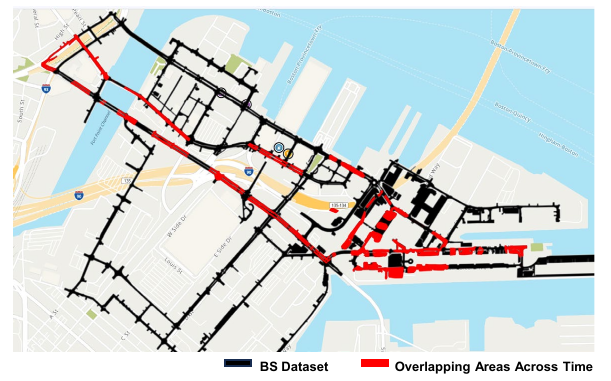}
  \caption{Visualization of the data distribution in the BS split of our proposed benchmark. The black areas represent the regions included in the BS split, while the red areas indicate segments where the same scene is revisited over time.}
  \label{fig:data}
  \vspace{-0.6cm}
\end{figure}

\section{Experiments}
\label{sec:experiments}
\subsection{Dataset and Benchmark}
\label{sec:data_setup}
Since our proposed method requires multi-view RGB images covering a $360^{\circ}$ field of view and LiDAR point clouds as inputs, we conduct experiments on the publicly available nuScenes dataset~\cite{caesar2020nuscenes}. The nuScenes dataset contains 1000 driving scenes, each with a duration of 20 seconds. A comprehensive sensor suite is equipped to collect multimodal data, including six cameras, a LiDAR, and five radars. It's collected at different locations, times, and weather conditions in Boston Seaport (BS), SG-OneNorth (SON), SG-Queenstown (SQ), and SG-HollandVillage (SHV). Among these, the BS, SON, and SQ splits have a rich variety of scenes and include multiple revisits to the same locations at different times, providing a basis for evaluating map update performance. To obtain statistically significant results, we conduct experiments on the BS, SON, and SQ splits.

To evaluate the effectiveness of our proposed LT-Gaussian, we introduce a Gaussian-map update benchmark based on the nuScenes dataset~\cite{caesar2020nuscenes}. ~\figref{fig:data} shows the visualization of the BS split data of our benchmark on the satellite image. Firstly, we decompose the nuScenes dataset into a set of sequences of length \(N=1+N_{\text{h}}+N_{\text{p}}\), which serves as the basic unit for scene reconstruction. Subsequently, sequences with timestamps smaller than \(ts_{\text{old}}\) are assigned to the \textit{old scene set}, and sequences with timestamps greater than \(ts_{\text{new}}\) are assigned to the \textit{new scene set}. We maximize \(N_{\text{day-interval}}=ts_{\text{new}}-ts_{\text{old}}\), subject to sufficient available sequences, to ensure a long time span between the old and new scenes. Lastly, after a distance-based downsampling, for each sequence in the \textit{old scene set}, we search for the closest sequence in the \textit{new scene set}, forming a scene pair. All available scene pairs are then compiled into the benchmark.

The details of the benchmark are shown in Table \ref{tab:data_org}. We aim to ensure that the scene pairs reflect common environmental changes that occur over a long time span, including structural changes, weather variations, and illumination changes. This allows for a comprehensive evaluation of the ability of our proposed method to update the Gaussian-map.

\begin{table}[t]
    \centering
    \renewcommand\arraystretch{1.1}
    \setlength{\tabcolsep}{8pt}
    \caption{Details of our map update benchmark}
    \label{tab:data_org}
    \begin{tabular}{cccc}
        \toprule
        Location & Boston Seaport & SG-OneNorth & SG-Queenstown\\
        \midrule
        $N_{\text{pairs}}$ & 312 & 201 & 114 \\
        $N_{\text{day-interval}}$ & 40 & 30 & 30 \\
        Time & Day & Day & Day+Night \\
        Weather & Rainy & Sunny & Sunny \\
        \bottomrule
    \end{tabular}
    \vspace{-0.6cm}
\end{table}

\subsection{Implementation Detail}
\label{sec:expe_setup}
When setting the benchmark, we set the number of historical frames relative to the sequence center to \(N_{\text{h}}=1\) and the number of future frames to \(N_{\text{p}}=1\). The values of \(ts_{\text{old}}\) and \(ts_{\text{new}}\) vary with the splits to balance the number of available scene pairs \(N_{\text{pairs}}\) with the time interval \(N_{\text{day-interval}}\). We generate the Gaussian-map of the old scene using Multimodal 3D-GS. To reduce the training time and regulate the number of Gaussians, we set the number of training iterations to 4000 and the densification interval to 100. The settings of the remaining parameters are the same as in vanilla 3D-GS~\cite{kerbl20233d}.

\begin{table*}[ht]
\centering
\begin{center}
\setlength{\tabcolsep}{11pt}
\renewcommand\arraystretch{1.2}
\caption{Comparison of scene reconstruction performance on the nuScenes Dataset}
\begin{tabular}{cccccccccc}
\toprule
\multirow{2}{*}{Methods} & \multicolumn{3}{c}{BS split} & \multicolumn{3}{c}{SON split} & \multicolumn{3}{c}{SQ split} \\ \cline{2-10}
 & SSIM$\uparrow$ & PSNR$\uparrow$ & LPIPS$\downarrow$ & SSIM$\uparrow$ & PSNR$\uparrow$ & LPIPS$\downarrow$ & SSIM$\uparrow$ & PSNR$\uparrow$ & LPIPS$\downarrow$ \\
\hline
Multimodal 3D-GS & 0.7315 & 18.4279 & 0.4289 & 0.6726 & 18.5902 & 0.4908 & 0.4768 & 21.5931 & 0.5476 \\
LT-Gaussian (ours) & \textbf{0.7391} & \textbf{19.2140} & \textbf{0.4160} & \textbf{0.6768} & \textbf{19.4760} & \textbf{0.4714} & \textbf{0.4791} & \textbf{23.1321} & \textbf{0.5236} \\ \hline
Improvement & 1.04\% & 4.26\% & 3.10\% & 0.62\% & 4.77\% & 3.95\% & 0.48\% & 7.13\% & 4.38\% \\
\bottomrule
\end{tabular}
\label{tab:nuScenes}
\end{center}
\vspace{-0.6cm}
\end{table*}

\begin{figure*}[ht]
  \centering
  \includegraphics[width=1\linewidth]{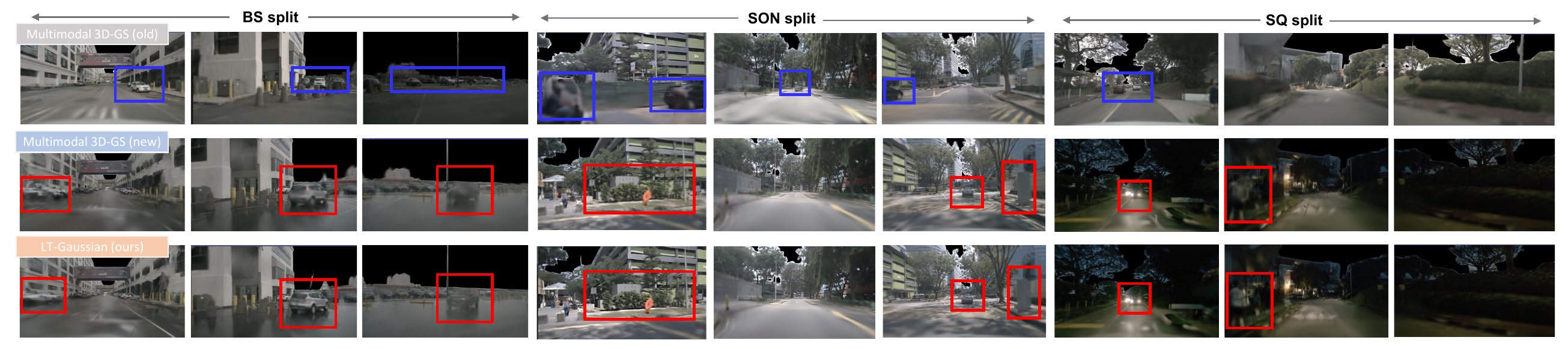}
  \caption{A comparison of our LT-Gaussian and Multimodal 3D-GS visualisation results. Representative ones from the BS, SON, and SQ datasets are listed, where blue boxes represent disappearing objects and red boxes indicate emerging objects.}
  \label{fig:5}
  \vspace{-0.6cm}
\end{figure*}

\begin{table}[ht]
\centering
\vspace{0.2cm}
\setlength{\tabcolsep}{12pt}
\renewcommand\arraystretch{1.2}
\caption{Map update time consumption comparison}
\begin{tabular}{cccc}
\toprule
\multirow{2}{*}{Methods} & \multicolumn{3}{c}{Time [s]} \\ \cline{2-4}
 & BS split & SON split & SQ split \\ \hline
Multimodal 3D-GS & 134.22 & 138.10 & 142.41 \\
LT-Gaussian (ours) & \textbf{38.64} & \textbf{40.12} & \textbf{41.28} \\ \hline
Improvement & 71.21\% & 70.95\% & 71.02\% \\
\bottomrule
\end{tabular}
\label{tab:time}
\vspace{-0.6cm}
\end{table}

After obtaining the Gaussian-map of the old scene, we perform ICP registration between the current LiDAR submap and the old Gaussian-map. During the registration process, we set the max correspondence distance \(d=2.0\,m\), and the max iterations \(n_{\text{iter}}=100\). 
Then we use KNN for environmental change detection, with the KNN detection threshold set to \(r=1.0\,m\) and \(r'=1.0\,m\). Through the detection results, we identify the emerging points and the disappearing points to obtain the Gaussian prior \(G_{\text{m}}\). In the process of generating the Gaussian-map for the new scene, we perform 1000 iterations of refinement with a densification interval of 200 based on the Gaussian prior \(G_{\text{m}}\). All experiments are conducted on a system equipped with an Intel i9-13900K CPU and an Nvidia RTX 4090 GPU.

\subsection{Quantitative Analysis}
\label{sec:eval_res}
In this section, we conduct experiments on our proposed map update benchmark. The goal is to demonstrate the superiority of our proposed LT-Gaussian in terms of both map update effectiveness and efficiency. We use a map update method that reconstructs environmental maps from scratch as the baseline. Specifically, we perform Multimodal 3D-GS-based reconstruction on data from the new sequence in the scene pairs, to obtain the Gaussian-map for the new scene. The LT-Gaussian, on the other hand, simulates the map update process based on structural change detection. We first construct the outdated Gaussian-map on data from the old sequence, then follow our proposed pipeline to perform the map update, ultimately obtaining the Gaussian-map for the new scene.

We report metrics, including Structural Similarity Index (SSIM), Peak Signal to Noise Ratio (PSNR), and Learned Perceptual Image Patch Similarity (LPIPS), to evaluate the quality of the updated Gaussian-map. These metrics reflect the similarity between the images rendered from the Gaussian-map and the ground truth images. The higher the similarity between the rendered image and the ground truth image, the better the updated Gaussian-map restores the geometric structure and encodes the texture information of the new scene. We randomly select 5\,\% of the images from each new sequence as the test set, with the remaining images used as the training set. We compute the mean SSIM, PSNR, and LPIPS values for the test set images and report them as statistically significant results. Furthermore, the average time consumption for map updates is reported as a measure of efficiency.

The experimental results are shown in Tab \ref{tab:nuScenes} and Tab \ref{tab:time}. Notably, our proposed LT-Gaussian method outperforms the baseline of reconstructing maps from scratch in terms of SSIM, PSNR, and LPIPS. This indicates that the maps obtained by our Gaussian-map update method effectively reflect the fine-grained geometric features, as well as semantic and texture information, in the new scenes. A possible reason is that our map update method not only effectively encodes the environmental features from the new scene, but also utilizes the features provided by the old scene, resulting in better reconstruction quality.

Furthermore, our method achieves faster map updates while improving the reconstruction quality. Our proposed method completes the map update training in approximately 27\,\% of the time required for reconstructing maps from scratch, demonstrating its efficiency.

\subsection{Visualization of Map Updates}
\label{sec:eval_range}
In this section, we present a qualitative comparison of the map updating results between our LT-Gaussian and the baseline of reconstructing maps from scratch. ~\figref{fig:5} visualizes several reconstructed scenes from the BS, SON, and SQ splits. As can be seen, our LT-Gaussian effectively updates the outdated Gaussian-map. The updated map accurately reflects the structural changes, while also encoding the variations in weather, illumination and texture of the new scene. In addition, the Gaussian-map of the new scene obtained by our method demonstrates higher quality compared to the baseline. By effectively utilizing the outdated map, LT-Gaussian produces more geometrically accurate reconstruction results. As a result, the rendered images show clearer details and more realistic visual effects.

\section{Conclusion}
\label{sec:conclusion}
In this paper, we present LT-Gaussian, a map update method for 3D-GS-based maps in autonomous driving scenarios. Our approach consists of three main components. Multimodal Gaussian Splatting integrates multi-view RGB images and LiDAR point clouds to generate Gaussian-map, serving as a rendering-quality map for autonomous driving. The Structural Change Detection module compares the outdated Gaussian-map with the current LiDAR data stream to identify structural changes. Based on this, the Gaussian-Map Update module performs targeted updates to the Gaussian-map. We establish a benchmark for map updating on the nuScenes dataset, and the experimental results are remarkable. LT-Gaussian outperforms the baseline, which reconstructs maps from scratch, achieving better reconstruction quality and significantly reducing map update time consumption.


\bibliographystyle{unsrt}

\footnotesize{
\bibliography{glorified, new}}

\end{document}